# 3D vision-based structural masonry damage detection

Elmira Faraji Zonouz[1*], Xiao Pan[2], Yu-Cheng Hsu[3], Tony Yang[4]

[1] MASc. candidate, Department of Civil Engineering, University of British Columbia, elmiraf@mail.ubc.ca
[2] Research fellow, Department of Civil Engineering, University of British Columbia
[3] PhD. student, Department of Civil Engineering, University of British Columbia
[4] Professor, Department of Civil Engineering, University of British Columbia

**ABSTRACT**

The detection of masonry damage is essential for preventing potentially disastrous outcomes. Manual inspection can, however, take a long time and be hazardous to human inspectors. Automation of the inspection process using novel computer vision and machine learning algorithms can be a more efficient and safe solution to prevent further deterioration of the masonry structures. Most existing 2D vision-based methods are limited to qualitative damage classification, 2D localization, and in-plane quantification. In this study, we present a 3D vision-based methodology for accurate masonry damage detection, which offers a more robust solution with a greater field of view, depth of vision, and the ability to detect failures in complex environments. First, images of the masonry specimens are collected to generate a 3D point cloud. Second, 3D point clouds processing methods are developed to evaluate the masonry damage. We demonstrate the effectiveness of our approach through experiments on structural masonry components. Our experiments showed the proposed system can effectively classify damage states and localize and quantify critical damage features. The result showed the proposed method can improve the level of autonomy during the inspection of masonry structures.

Keywords: Computer Vision, Deep Learning, Masonry Structures, Automatic Damage Detection, Structural Health Monitoring

**INTRODUCTION**

The durability and strength of masonry buildings make them a popular choice for construction worldwide. Masonry structures have many different styles, from traditional to modern, and are used in a wide range of applications, including residential and commercial. A masonry structure typically consists of a range of materials such as bricks, ashlars, blocks, and stones, which are commonly held together using a mortar. Spalling, cracks, efflorescence, and discoloration are types of masonry damage as well as rebar exposure in reinforced modern masonry structures.

The inspection and maintenance of masonry architecture traditionally depend on tactile examination, visual inspection, wave transmission, ultrasonic velocity testing, and hammer tests [1-2]. However, these techniques have limitations, as they tend to be time-consuming, manual, destructive, and may lead to high workloads. Additionally, there may be potential inaccuracies due to lack of experience or engineer bias, and the techniques may pose risks to inspectors. In recent years, new technologies and devices, such as unmanned aerial vehicles (UAVs), unmanned ground vehicles (UGVs), Light Detection and Ranging (LiDAR), and photogrammetry have emerged as potential solutions to these challenges, facilitating faster and more accurate inspection procedures without causing damage [2-3].

The implementation of computer vision technology has revolutionized the process of identifying and localizing damage in various types of structures and infrastructure [7]. This methodology employs visual inspection devices to capture RGB images or 3D point clouds, which are then processed using advanced machine learning (ML) and deep learning (DL) techniques. The resulting output is an accurate segmentation and localization of damage in a quick and efficient manner. The efficacy of this approach has been demonstrated in several scenarios, including the monitoring of bridges, concrete buildings, heritage masonry, and other structures [4-5]. Notably, this methodology is a significant improvement over traditional techniques, which are time-consuming and require extensive labor.

Computer vision technology is primarily utilized to detect damage, cracks, spalling, delamination, efflorescence, and rebar exposure in concrete and masonry structures [3, 6-9]. Numerous studies have focused on detecting and localizing damage in structures by analyzing 2D images of concrete columns, bridges, walls, and masonry heritage buildings and bridges [10-12]. To this end, Convolutional Neural Networks (CNNs) such as YOLO, R-CNN, Faster R-CNN, and MobileNet have been widely employed for both object detection and image classification [11, 13-15]. While object detection involves localizing damage types through bounding boxes, image classification assigns a label or category to an entire image, and both tasks rely on the capacity of CNNs to learn and extract meaningful features from raw image data.

In addition to object detection and image classification, another commonly used computer vision task is semantic segmentation, which involves labeling each pixel in an image with a class label. Although the aforementioned YOLO, R-CNN, Faster R-CNN, and MobileNet models are primarily designed for object detection and classification [10, 13], they can also be adapted for semantic segmentation tasks by incorporating additional layers or modifying the output format. Nonetheless, there are other CNN-based models specifically designed for semantic segmentation, such as U-Net and SegNet. CNN-based methods for structural damage inspection have been widely investigated for various civil infrastructure [8,13,16-20, 50]. Some techniques have been previously used to distinguish structural components, such as bricks or stones in masonry walls, or segmenting damages, such as cracks or spalling, mostly in concrete or masonry surfaces [21-24].

On the other hand, Transformer-based models were originally designed for natural language processing tasks but have recently been applied to computer vision tasks such as object detection and image segmentation. Some studies have shown that transformer-based models can outperform most existing models for surface defect detection in industrial applications [25]. Moreover, hybrid encoders combining CNN-based and transformer-based encoders, such as TransUNet-hybrid, have shown superior accuracy, generalization, and robustness in crack-detection models in masonry structures compared to CNN-based models [26].

Although damage detection in structures has been largely based on 2D image-based methods using CNNs and Transformer-based models, these methods have some limitations. For instance, the accuracy of these models relies heavily on the lighting conditions, which can pose a challenge in real-world scenarios. Moreover, these models may require significant computation time and training data to achieve an optimal performance. Another drawback is the lack of quantification in most cases, as they only detect the presence of damage, without providing detailed information on the measurement.

By contrast, 3D point-cloud-based methods offer several advantages for damage detection in structures. One significant advantage is that they can provide a more comprehensive representation of a structure's condition by capturing its geometry and spatial arrangement. Additionally, 3D point clouds are less affected by lighting conditions because they rely on geometric information rather than image intensity, especially when LiDAR or laser scanners are used to create point cloud data (PCD). These methods also offer more opportunities for quantification, as they can provide detailed information on the geometry, specifically the depth and location of damage, allowing for a more accurate assessment of their severity and extent.

Recently, some studies have employed 3D point cloud data mainly created by laser scanners or LiDAR devices, and the structure from motion (SFM) technique to detect or segment elements or defections in buildings, bridges, and pipelines [27-30]. Additional processing and analysis are required to extract useful information from point cloud data. The analysis methods for the point cloud data (PCDs) of structural components can be categorized into two groups: those based on artificial intelligence (AI), including machine learning (ML) and deep learning (DL) techniques [28], and those that are not AI-based, such as [31-32]. Both approaches have their own advantages and drawbacks, and choosing the appropriate method depends on the specific task and the available resources.

AI-based methods have mostly been used to segment components [33-35] or damage reinforced concrete bridges [29, 36-38]. These studies typically involve creating a dataset of annotated PCDs for elements such as slab, girder, pier, cap pier, abutments, girders, or defects (e.g., cracks or spalling), training well-known DL methods such as PointNet [34-35] or DGCNN [35], or proposed algorithms such as SNEPointNet++ [29], to automatically distinguish between components or damages. Similar DL-based algorithms such as MVCNN, PointNet, PointNet, PointNet++, DGCNN, PCNN, and modified DGCNN have been applied to segment building components such as column, floor, wall, roof, door, stairs etc. in [27, 39]. Many structural applications lack the labeled data required to use DL-based methods. In addition, training a satisfactory classification or segmentation algorithm requires time and computational costs, even with adequate data [40].

One potential solution to address training issues is the use of unsupervised learning techniques. For instance, unsupervised clustering methods, including k-means, fuzzy c-means, SC, and DBSCAN, have been employed to segment corrosion, erosion, and spalled off tiles on rolling doors or exterior walls [41]. In another application, an unsupervised anomaly classification technique, known as OC-SVM, was utilized for RC bridge defect classification [37].

The amount of data required for unsupervised algorithms depends on the specific application and the complexity of the dataset. However, the need for sufficient data to achieve accurate clustering results can persist in some applications. Also, sensitivity to parameters and initial values, local minimum convergence, and usage complexity can still be drawbacks of using clustering methods [42]. Although some studies have already been conducted, there are many fields that can be investigated or improved. Moreover, limited research has been conducted on directly quantifying the damage in the PCDs of concrete surfaces [31, 43].

In summary, the use of 3D point cloud data has become increasingly popular in various fields for detecting and segmenting structural elements and defects. Point cloud data require additional processing and analysis, which can be accomplished either using AI-based methods or non-AI-based methods. Although AI-based methods have shown great promise in segmenting components or damages, they often require labeled data and can be computationally expensive. A potential solution to these issues may be clustering algorithms, which are sensitive to parameters and initial values. Therefore, much remains to be explored and improved in this field.

In this study, we propose a new approach for segmenting spalled regions in concrete masonry blocks using geometric primitives in point cloud data. Our method is fast and does not require large amounts of data or training algorithms, making it highly efficient for practical applications. Previous studies on automatic damage detection in masonry have mainly focused on heritage buildings, with little attention paid to modern reinforced masonry elements. As modern masonry structures have become more prevalent, it has become increasingly important to be able to inspect and maintain them during their lifetime. However, traditional methods have proven problematic; therefore, it is necessary to apply new methods to improve the process. Previous studies have primarily used RGB images for segmentation which is limited to evaluation in 2D space, thus unable to provide detailed and accurate quantification in 3D space. Our approach, instead, employs 3D vision method to reconstruct the 3D geometry of masonry blocks, which will be analyzed to quantify their damage features accurately in 3D space.

To implement our approach, we first created 3D point clouds using the structure-from-motion (SFM) method with a cheap and accessible phone camera. We then automatically removed outlier points and applied planar patch detection and plane segmentation to quickly detect the faces of the blocks and subsequently identify the dent areas on the selected surfaces. Finally, the spalled volume was estimated using the convex hull method. Overall, our method offers a promising solution for efficient and accurate damage detection and volume estimation of modern masonry structures.

The rest of this paper is organized as follows. Section 2 provides a description of our methodology, including an approach to creating point cloud data from the images in Subsection 2.1. Subsection 2.2 describes our method for assessing point cloud data to segment spalling damage in concrete blocks. In Subsection 2.3, we present a mathematical approach to quantify the damage. Finally, Section 3 draws conclusions and discusses future directions for this field of research.

**METHODOLOGY**

**The procedure of 3D reconstruction**

There are several techniques and devices that can be used to create a point cloud data (PCD) from a specific object, such as laser scanners, LiDAR, and structure from motion. This paper focuses on using the structure from motion (SFM) method to create point cloud data. While SFM may take longer than other methods to create a 3D point cloud, it is a low-cost method, especially when LiDAR or laser scanners are not available. This method can be done through DSLR or even phone cameras, which are widely accessible today.

SFM is a computer vision technique that uses sequential 2D images to create a 3D model of a scene or object [44]. To create a 3D reconstruction by SFM from an object, the object should be inanimate, non-reflective, and accessible from 360 degrees to create a complete 3D reconstruction. SFM involves detecting and matching feature points, estimating camera poses, and triangulating points to obtain a 3D model. SFM has many practical applications, including photogrammetry, 3D mapping, augmented reality, and robotics.

Although using terrestrial laser scanners (TLS) or UAVs equipped with LiDAR is preferred for creating PCDs of large infrastructures or structures, SFM can effectively create 3D points from structural elements or indoor building objects. Therefore, SFM was used in this paper since the objective structural elements are masonry concrete blocks and prisms, which are small. So, the SFM method is reasonably fast, accurate, and cheap for these elements. Additionally, when high-resolution images are used for SFM, the 3D reconstructed model created by this method has a higher accuracy in showing even small dents. All these factors make the SFM method highly applicable for our aim.

This study utilized approximately 20-30 images taken from different angles using a Samsung Galaxy S20 FE to create a 3D reconstruction of each object. Then, Autodesk Recap Photo student version (v.2023) was used to generate 3D models.

Figure 1 shows various 3D reconstructed models of masonry blocks and prisms. Specimens were photographed under both natural and artificial lighting conditions before and after tests. Image resolution was $4032 \times 1816$, with a lens aperture of F1.8 and a focal length of 26 mm. Shutter speed and ISO were adjusted based on light conditions. It is possible to capture even the smallest dent in these images due to their high level of detail.

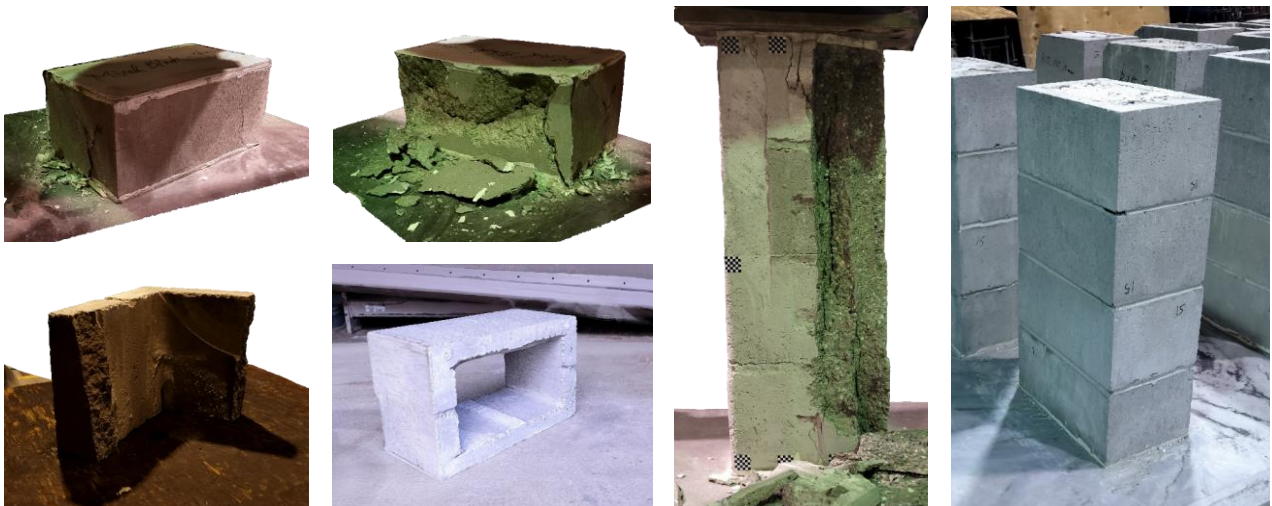

*Figure 1. Samples of 3D reconstructed models of masonry blocks and prism before and after compression test under indoor or outdoor lighting conditions.*

**Assessment of 3D point cloud data**

In this section, the reconstructed models were analyzed due to their ability to contain both geometrical (x, y, z locations) and color information (RGB) in the form of a 3D point cloud. Then, geometrical algorithms were used to segment the spalled region in those 3D reconstructed models of concrete blocks subjected to the compression test done by the Baldwin machine.

First, the noises should be removed from the original object. This procedure can be done manually or automatically by outlier remover algorithms with the help of Open3D library [45] which is used in this paper for further processing of PCDs. Open3D is a free library that supports 3D data. Both C++ and Python are used for the frontend, which exposes data structures and algorithms carefully selected for their performance and utility. Parallelization is built into the backend, which is highly optimized. Open3D library is employed in the next part of this study to visualize and process point cloud data.

In Figure 2, a 3D point cloud sample obtained from a concrete block after eliminating excessive noise is observed. The dimensions of the block are approximately 390x240x190 mm3, and the sample contains a total of 127664 points. Notably, this study relies solely on geometrical features to identify spalling segments, and the color displayed in the figure corresponds to the coordination intensities from the origin.

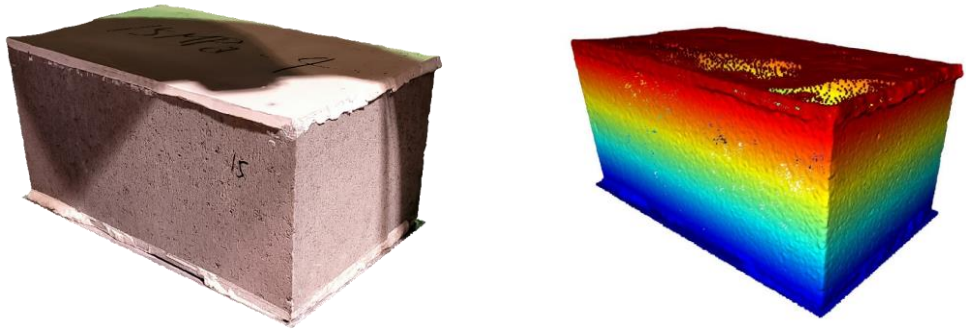

*Figure 2. 3D reconstructed model versus 3D point cloud.*

To effectively isolate spalling areas from the original shape, a novel approach was employed that combines both plane segmentation and planar patch detection. To achieve this, a custom Python code was developed that can detect individual faces within each block. Once a face has been identified, a plane is then segmented to represent the original face of the block. Notably, any spalled portions of the block face are specifically excluded from the plane, as they do not conform to the original plane shape of the block.

To begin with, a "planar patches detection" algorithm, as introduced by [46], is employed to detect each face of the concrete block specimen. This algorithm is designed to identify planar patches in the point cloud, which are regions that lie on the same plane. The reason for choosing this algorithm is that it can be applied to various applications such as object recognition, reconstruction, and segmentation, which also aligns with our segmentation goal.

The algorithm as a robust statistics-based approach, initially breaks down the point cloud into smaller, more manageable chunks. From there, it tries to fit a plane to each of these smaller segments. This approach has proven to be highly effective in accurately identifying planar patches within complex point clouds. After finding each face of a block, then each patch saves as an input for plane segmentation in the next step to detect spalled region.

First, the code load then read the input dataset includes PCD as PLY file. Then the algorithm check if the point cloud has normal vectors, which are important for planar patch detection. The detection process can be controlled using various parameters. These parameters control the variance of the surface normal vectors, the degree of coplanarity between points, and the ratio of outlier points allowed in the patch.

The plane segmentation from PCDs a fundamental task in computer vision and robotics. This process involves identifying and separating planar surfaces from a collection of data points, which can be useful for various applications such as object recognition, reconstruction, or navigation. The aim is to estimate a robust model from noisy data, which can accurately represent the underlying geometry of the environment.

The code presented here demonstrates the use of the RANSAC algorithm [47] to perform plane segmentation on a point cloud. RANSAC, which stands for "Random Sample Consensus," is an iterative algorithm that is proper for fitting models to data points that contain noise or outliers. The algorithm works by randomly selecting subsection of data points and fitting models to these subsets. The fitted models are then evaluated on the remaining data points, and if the model fits well enough to a certain threshold, it is considered a "consensus" model.

In this code, the input point cloud data is loaded from a PLY file, and then the RANSAC algorithm is applied to fit a plane to the input point cloud. The algorithm returns the plane parameters and inlier indices, which can be used to extract the planar region from the original point cloud. This process is repeated multiple times, and the best consensus model is selected as the final model.

The segmentation process is sensitive to three control parameters, namely the minimum number of points, distance threshold, and number of iterations [48]. These parameters control the accuracy and sensitivity of the segmentation, and their values were chosen based on trial and error to optimize the performance for the specimens.

In our study, the algorithm initially selects a minimum of three points randomly from the input point cloud data to determine the model parameters. Subsequently, the code evaluates how many points from the entire set of points fit within a predefined tolerance, which was set to $\tau=0.002$ for our objects. If the fraction of inliers over the total number of points in the set exceeds a predefined threshold, the algorithm estimates the model parameters again using all the identified inliers and terminates. However, if the threshold is not met, the algorithm repeats the same steps for a maximum number of iterations, which is set to 100 for our application. It is important to note that the control parameters were chosen based on the specific requirements of the first block and were subsequently applied to other specimens. The function finally returns the plane equation based on (x, y, z) coordination, and further returns a list of indices of the inlier points.

Based on the segmentation results, the code extracts and visualizes the inliers and outliers from the input point cloud. A selection by indexes is used to filter the point cloud based on the inlier indices. To visually distinguish the spalled region from the rest of the block, it has been colored red. On the other hand, the original plane, and the points which can be considered as the outer part of the plane, has been colored black to differentiate them from the spalled region.

Figure 3 displays the spalled region within a block as a sample of code output. The damaged region is depicted in red, and it can be observed that it has been effectively segmented from the original specimen. that the shape depicted in Figure 3 represents one-fifth of a block face, allowing for a more detailed illustration. As it observed, the spalled area has been accurately identified and distinguished from the rest of the block.

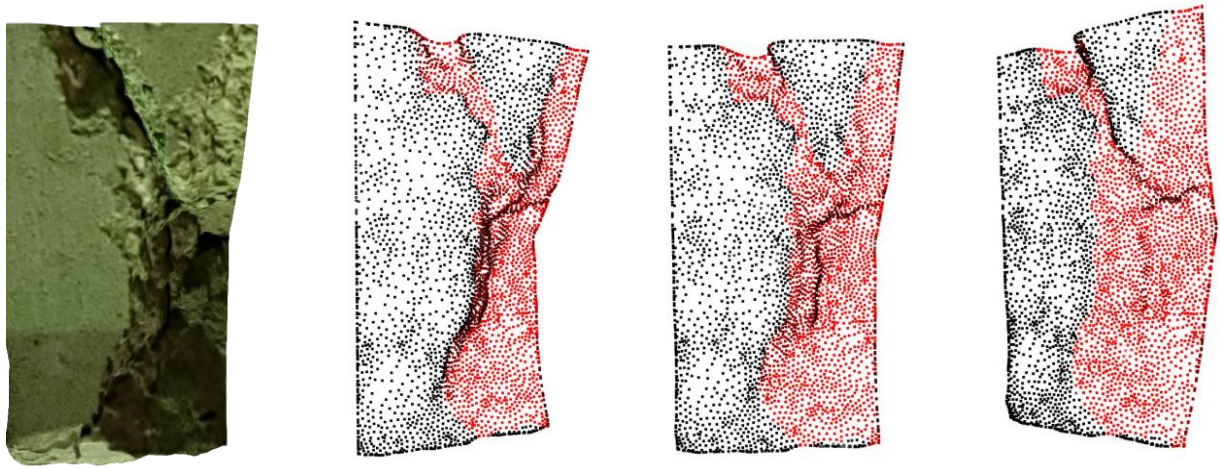

*Figure 3. Segmented spalled region in PCD of a concrete block under compression test versus the 3D reconstructed model.*

**Quantifying damage**

Only a limited amount of research has been conducted on directly quantifying damage in PCDs, with only a handful of studies validating the use of convex hull algorithms [43] to measure spalling on concrete surfaces. In this study, we employed a fast and powerful mathematical tool for the convex hull algorithm to measure the spalled concrete volume on each damaged face of the specimen. The convex hull algorithm is particularly useful for fitting a convex polyhedron that closely approximates the shape of the damaged volume because it collects all points in the smallest set of convex shapes in a point cloud [49]. To obtain the exact spalled volume, scaling of the object size should be performed, which requires knowing the real dimensions of the specimen. We first verified the effectiveness of this method by measuring an undamaged concrete block with known dimensions, and then utilized it to quantify the spalled region (colored in red in Figure 3). Figure 4 shows a convex hull containing the spalled region of the chosen concrete face. This algorithm estimated the spalled volume as 0.0005 m3 (500,000 mm3), which is equivalent to 2.8% of the entire block. Overall, the convex hull algorithm is a simple, efficient, and accurate method for effectively estimating damaged volume.

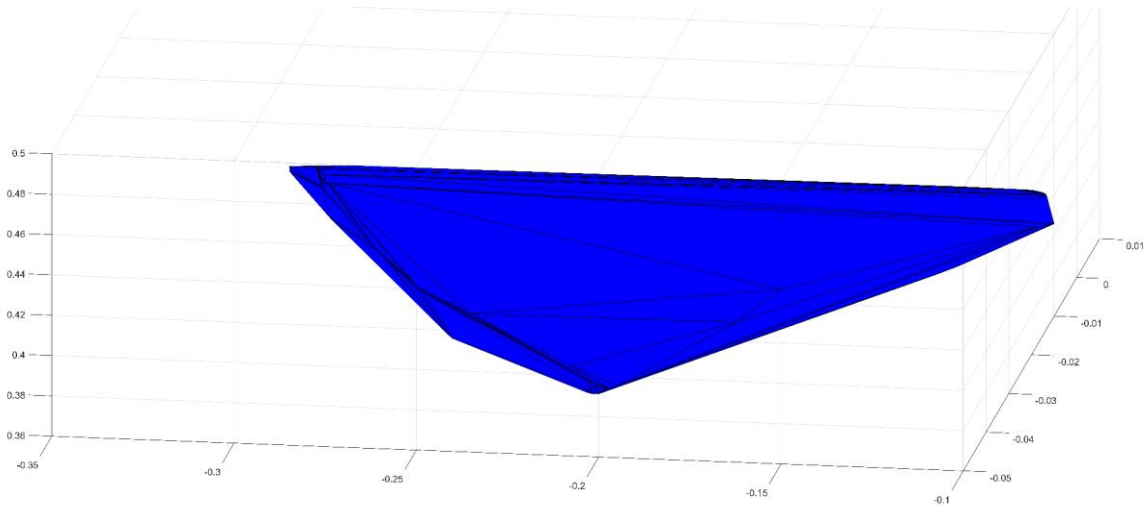

*Figure 4. An overview of the convex hull estimating the spalled concrete volume in a specimen.*

## CONCLUSIONS

This study presents a novel approach for segmenting spalled regions in modern masonry structures using geometric primitives in point cloud data created using the SFM method applied to multiple images of a specimen. The proposed method is fast and accurate and does not require large amounts of data or training algorithms. Unlike previous studies, which mainly focused on heritage buildings, this research demonstrates the potential of applying 3D point cloud data for the efficient and accurate segmentation of spalled regions in modern masonry components.

The results showed that the proposed method could effectively localize spalling damage and quantify the segmented damaged volume of the masonry block. This method is an efficient way to segment the damaged region, but we lack sufficient data or time to train a deep learning model.

However, the speed of this method could be improved by applying LiDAR to accelerate the process of creating 3D point clouds and removing the need for scaling to the original geometry of the object. Therefore, measurements can be performed without knowing the dimensions of the structure. Changing its behavior from a semi-manual to an automatic process can improve performance.

In this study, only concrete block specimens were investigated, and it is crucial to apply the method to larger masonry elements, such as walls, which will be considered in the next steps following this study by improving the method. Overall, this study offers a promising solution for the efficient and accurate damage detection and volume estimation of modern masonry structures and can assist in the inspection and maintenance of such structures during their lives.

## ACKNOWLEDGMENTS

We would like to acknowledge the support received from the NSERC Alliance Grant as well as our industry partners, the Canada Masonry Design Centre (CMDC), and the Canadian Concrete Masonry Producers Association (CCMPA). Their generous support and contributions were instrumental to the success of this research project.